\pdfoutput=1

\documentclass[11pt]{article}

\usepackage{amsmath,amsfonts,bm}

\def\eqref#1{equation~\ref{#1}}

\def\1{\bm{1}}

\DeclareMathAlphabet{\mathsfit}{\encodingdefault}{\sfdefault}{m}{sl}
\SetMathAlphabet{\mathsfit}{bold}{\encodingdefault}{\sfdefault}{bx}{n}

\def\gL{{\mathcal{L}}}

\def\gX{{\mathcal{X}}}
\def\gY{{\mathcal{Y}}}

\usepackage[final]{acl}
\usepackage{multirow}
\usepackage{times}
\usepackage{latexsym}
\usepackage{cleveref}
\usepackage[T1]{fontenc}
\usepackage{pifont}

\usepackage[utf8]{inputenc}

\usepackage{microtype}

\usepackage{inconsolata}

\usepackage{graphicx}
\usepackage{booktabs}
\usepackage{adjustbox}
\usepackage{colortbl}
\usepackage{xcolor}

\usepackage{subcaption}

\newcommand{\cmark}{\ding{51}}%
\definecolor{lightblue}{RGB}{173, 216, 230}

\title{Automatic Expert Discovery in LLM Upcycling via Sparse Interpolated Mixture-of-Experts}

\author{
  Shengzhuang Chen \\
  Thomson Reuters \\
  Foundational Research \\
\texttt{shengzhuang.chen@}\\\texttt{thomsonreuters.com} \\\And
  \text{Ying Wei$^{*}$} \\
  Zhejiang University \\
  \texttt{ying.wei@zju.edu.cn} \\ \\\And
  \text{Jonathan Richard Schwarz\thanks{Corresponding authors; Equal contribution.}} \\
  Thomson Reuters \\
  Foundational Research \\
  \texttt{jonathan.schwarz@}\\\texttt{thomsonreuters.com}
}

\begin{document}
\maketitle
\begin{abstract}
We present Sparse Interpolated Mixture-of-Experts~(SIMoE) instruction-tuning, an end-to-end algorithm designed to fine-tune a dense pre-trained Large Language Model (LLM) into a MoE-style model that possesses capabilities in multiple specialized domains. During instruction-tuning, SIMoE automatically identifies multiple specialized experts under a specified sparsity constraint, with each expert representing a structurally sparse subset of the seed LLM's parameters that correspond to domain-specific knowledge within the data. SIMoE simultaneously learns an input-dependent expert merging strategy via a router network, leveraging rich cross-expert knowledge for superior downstream generalization that surpasses existing baselines. Empirically, SIMoE consistently achieves state-of-the-art performance on common instruction-tuning benchmarks while maintaining an optimal performance-compute trade-off compared to all baselines.
\end{abstract}

\section{Introduction}\label{sec:introduction}
The rapid advancement of large language models (LLMs)~\citep{grattafiori2024llama3herdmodels, openai2024gpt4technicalreport} has revolutionized natural language processing, cementing their role as foundational tools across disciplines such as engineering~\citep{Hou2024LLMForSoftwareEng}, mathematics~\citep{RomeraParedes2023MathematicalDF}, humanities~\citep{ziems-etal-2024-large}, and the life sciences~\citep{Lin2022EvolutionaryscalePO}. While pre-trained LLMs demonstrate impressive general-purpose capabilities, their adaptation to specialized tasks often demands extensive instruction-tuning~\citep{Zhang2023InstructionTF}. This process typically involves supervised fine-tuning on domain-specific instruction datasets to align outputs with task requirements. Recent research highlights the critical interplay between scaling instruction-tuning data diversity, volume, and model capacity to achieve robust generalization~\citep{wei2022finetuned,Longpre2023TheFC,grattafiori2024llama3herdmodels}. However, balancing these dimensions efficiently without incurring prohibitive computational costs remains an open challenge.\par

This tension has spurred growing interest in Sparse Mixture-of-Experts (SMoE) architectures~\citep{shazeer2017}, which offer a promising pathway to flexible scaling of model capacity while maintaining inference efficiency by dynamically activating subsets of parameters per input for both training and inference. Yet, practical adoption of SMoEs in instruction fine-tuning is hindered by two key barriers: (i) the scarcity of publicly available pre-trained SMoE checkpoints~\citep{muennighoff2025olmoe} and (ii) the immense computational cost of (pre-)training them from scratch. To overcome these challenges, researchers and practitioners are increasingly exploring a cost-effective alternative known as \emph{upcycling}~\citep{komatsuzaki2023sparse}, which expands pre-trained dense LLMs into SMoE architectures by replacing subsets of their feed-forward networks (FFN) with SMoE modules. Although cost-effective, existing upcycling methods remain fraught with limitations that undermine their efficacy.\par

\emph{First}, manual selection of pre-trained parameters for upcycling: Existing methods upcycle FFN or attention blocks in pre-trained LLMs, assuming uniform utility across all upcycled layers. However, this fixed upcycling strategy fails to account for critical factors: (1) model-specific dynamics -- layers and parameters within the same pre-trained LLM can exhibit diverse properties and varying importance to model functionality, and (2) the highly domain-specific nature of instruction-tuning data, which may require different parts of pre-trained models to be upcycled and fine-tuned optimally. This disconnect between the algorithm, model, and data renders existing upcycling heuristics highly inflexible, preventing them from adaptively meeting the needs of specific instruction-tuning scenarios. As a result, existing upcycled approaches lead to suboptimal performance and diminish the generalization capabilities of upcycled LLMs. The benefits of dynamically identifying rather than manually specifying components are well documented \citep[e.g.]{von2021learning, schwarz2021powerpropagation, chen2024unleashing}\par 

\emph{Second}, a lack of systematic mechanisms to encourage expert specialization and cooperation. While ~\citet{dai2024deepseekmoe} proposes an architectural approach to promote expert specialization by using a fixed shared expert that is always active in addition to routed experts, recent work in~\citet{muennighoff2025olmoe} reports performance degradation with this approach, possibly due to limited adaptability with rigid expert partitions. Alternatively, some upcycling approaches explicitly promote specialization by fine-tuning domain-specific experts independently and then merging them into a unified SMoE architecture. However, these methods critically depend on high-quality domain labels and optimal domain partition -- requirements that are not readily met in practice. Consequently, upcycled experts often exhibit redundancy and fragmented specialization, undermining their efficacy.\par

\begin{figure}[t!]
\centering
\includegraphics[width=1.0\linewidth]{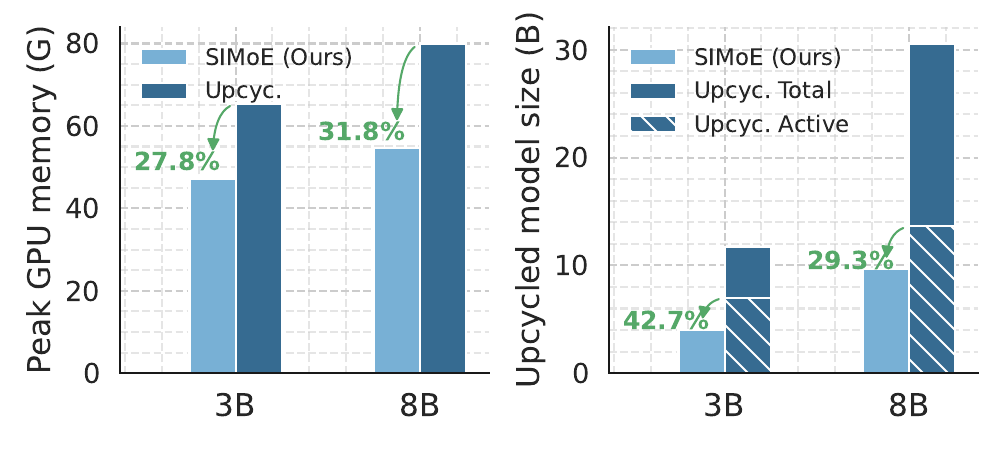}
\caption{Compute cost in terms of (\textit{left}) peak memory per GPU during upcycling instruction-tuning of 3B and 8B pre-trained LLMs, and (\textit{right}) number of model parameters during inference.}
\label{fig:train_test_compute}
\end{figure}

\begin{figure}[t!]
\centering
\includegraphics[width=1.0\linewidth]{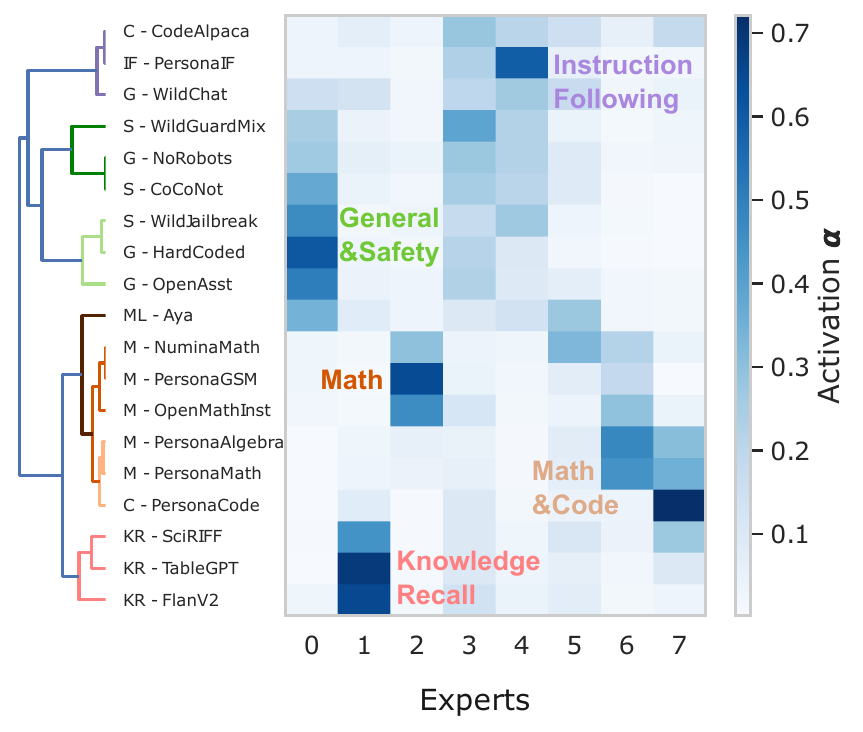}
\caption{Dendrogram illustrating task similarity
derived from learned expert activations, with prefixes indicating domain categories.}
\label{fig:dendrogram}
\end{figure}

\begin{figure*}[h!]
  \begin{center}
    \includegraphics[width=1.\textwidth]{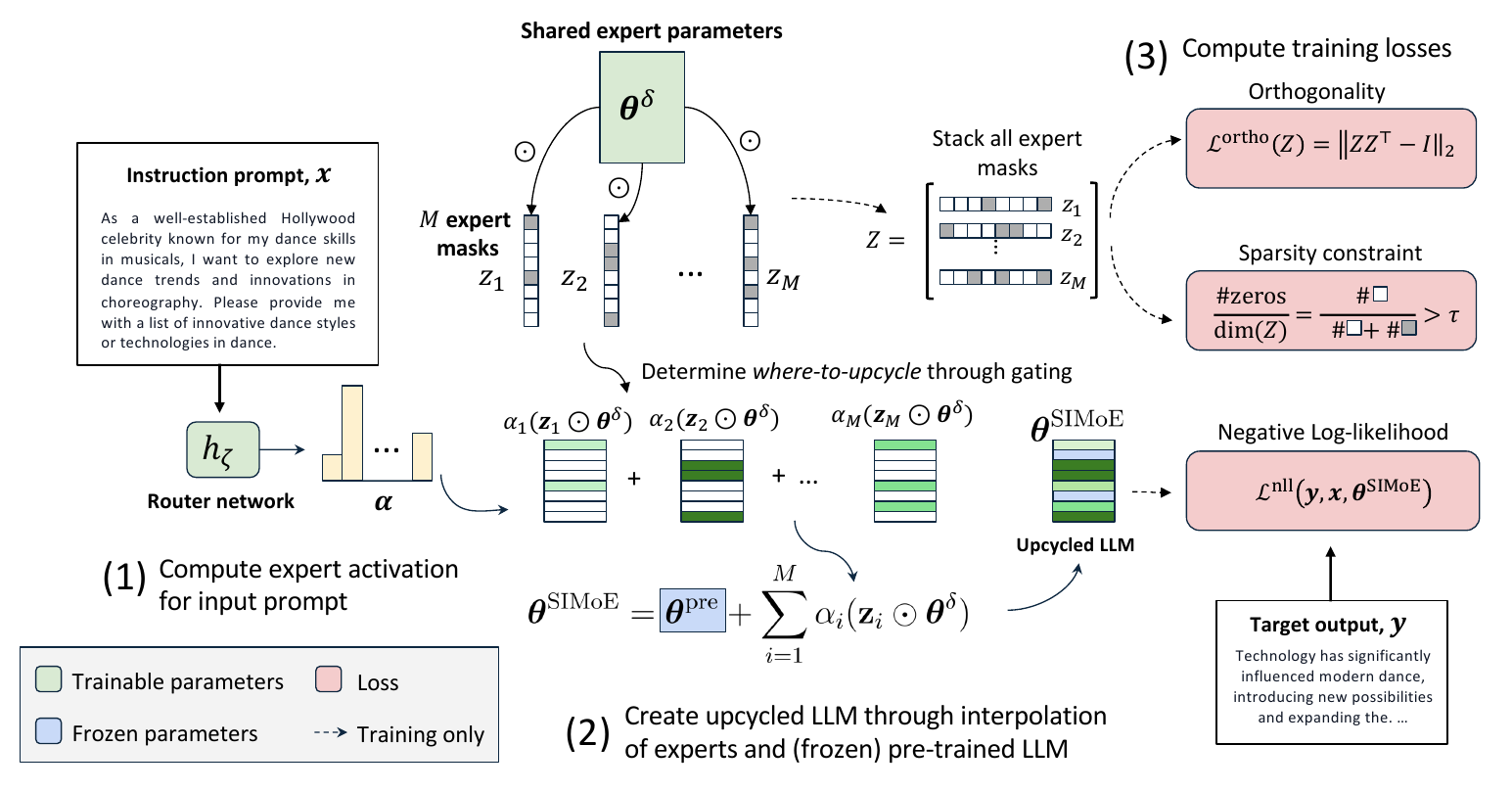}
  \end{center}
  \caption{\label{fig:overview}Overview of the proposed \textbf{S}parse \textbf{I}nterpolated \textbf{M}ixture-\textbf{o}f-\textbf{E}xperts~(SIMoE) instruction-tuning approach. SIMoE conceptually resembles the MoE principle in routing and combining specialized parameter components through soft merging, while it differs in implementation from conventional MoE architectures by defining each expert as a specific subset of sparse parameters within a shared network. Specifically, SIMoE upcycles a pre-trained LLM into a MoE-style model characterized by $M$ experts, consisting of a \textbf{shared}, trainable set of expert parameters $\boldsymbol{\theta}^\delta$ and $M$ \textbf{distinct}, trainable sets of expert masks $\{\boldsymbol{z}_m\}^{M}_{m=1}$. In forward computation, \textbf{(1-2)} SIMoE merges experts via a weighted-sum with coefficients $\boldsymbol{\alpha}_i$ generated via a router network $h_{\boldsymbol{\zeta}}$ based on the input prompt $\boldsymbol{x}$, before combining with the frozen, pre-trained LLM. \textbf{(3)} During instruction-tuning, we enforce structured sparsity and orthogonality on the trainable masks in addition to the usual NLL loss, determining \emph{where-to-upcycle} and encouraging expert specialization in a data-driven, fully automatic manner.}
  
\end{figure*}

These shortcomings point to an urgent need for an adaptive and automated framework that optimizes \emph{where-to-upcycle} and fosters expert specialization and synergy. To this end, we introduce \textbf{S}parse \textbf{I}nterpolated \textbf{M}ixture-\textbf{o}f-\textbf{E}xperts~(SIMoE), a novel algorithm and MoE architecture designed to address both challenges in upcycling instruction-tuning~(Fig.~\ref{fig:overview}). During training, SIMoE automatically identifies multiple experts for upcycling through sparsity-constrained optimization. Each expert represents a structurally sparse subset of the base LLM’s parameters, corresponding to specialized knowledge within the training data~(Fig.~\ref{fig:dendrogram}). Crucially, SIMoE enables learnable, soft parameter sharing between experts while imposing an orthogonal penalty to encourage specialization, thereby dynamically discovering a nuanced balance between synergy and specialization in the upcycled experts through optimization~(Fig.~\ref{fig:overlaps}). Our complementary innovations lead to empirical superiority in our method over strong baselines, delivering significant improvement of over 1.6\% in cross-task generalization~(Tab.~\ref{tab:fullft_sni}) with approximately $30$\% savings in training and inference memory costs~(Fig.~\ref{fig:train_test_compute}). \par
\newblock

\noindent{We summarize our main contributions as follows:}
\begin{itemize}
    \item We propose SIMoE, an effective and flexible method that systematically determines \emph{where-to-upcycle} and fosters expert specialization and synergy during instruction-tuning.
    \item We offer an effective solution that maintains scalability, enabling its application to pre-trained LLMs with billion-scale parameters.
    \item We empirically validate the efficacy of our method, demonstrating superior cross-task generalization performance on the Super-NaturalInstruction benchmark, as well as outperforming the recently open-sourced state-of-the-art $\text{T\"ulu}$-v3-8B-SFT~\citep{lambert2025tulu3pushingfrontiers} on common LLM benchmarks.
\end{itemize}

\section{Related Work}

\subsection{Sparse Mixture-of-Experts}  
Sparse Mixture-of-Experts (SMoE) architectures~\citep{jiang2024mixtralexperts,dai2024deepseekmoe, muennighoff2025olmoe} have gained prominence in both LLM pre-training and post-training due to their superior scalability compared to their dense counterparts. A standard SMoE architecture replaces the FFN $f: \mathcal{X} \rightarrow \mathcal{Y}$, which maps input to output, in the Transformer block~\citep{ashish2017attention} with a MoE module consisting of two components: (1) a set of $M$ experts $\{f_1, f_2, \ldots, f_M\}$, and (2) a router function $h: \mathcal{X} \rightarrow \mathbb{R}^M$ that outputs expert activations $\boldsymbol{\alpha} = [\alpha_1, \alpha_2, \ldots, \alpha_M]$ for input $\boldsymbol{x}$. The SMoE output $\boldsymbol{y}$ is then given by a weighted combination of expert outputs:  
\begin{equation}\label{eq:smoe_layer}  
    \boldsymbol{y} = \sum_{i=1}^{M} \alpha_i \cdot f_i(\boldsymbol{x}), \quad \text{where } \boldsymbol{\alpha} = h(\boldsymbol{x}).  
\end{equation}  

A defining feature of SMoE is its sparsity in expert activations: typically, only a subset of the most relevant experts is activated per input using the Top-$K$~\citep{Fedus2021SwitchTS} or Top-$P$~\citep{huang-etal-2024-harder} routing schemes. Early work has explored discrete expert-to-token routing -- selecting only the most activated subsets of experts for each token in $\boldsymbol{x}$~\cite{shazeer2017,Riquelme2021ScalingVW,lepikhin2021gshard, Fedus2021SwitchTS}, while later work introduced alternatives, such as token-to-expert routing -- choosing the top-scored subsets of tokens for each expert~\citep{zhou2022mixtureofexperts}, random and hash-based routing~\citep{Roller2021HashLF,zuo2022taming}. Advanced approaches further optimize routing stability and improve routing load-balancing~\citep{Lewis2021BASELS,liu2023sparsityconstrained}. \par

However, discrete and sparse expert activations often lead to training instability for gradient-based optimization~\citep{mustafa2022multimodal,dai-etal-2022-stablemoe}. To alleviate this issue, recent innovations propose soft, continuous routing alternatives. For example, SoftMoE~\citep{Puigcerver2023FromST} relaxes discrete assignments via continuous approximations, while SoftMerging \citep{Muqeeth2023SoftMO} explores combining experts through a weighted sum in the parameter space.\par

\subsection{Upcycling Dense LLMs to SMoEs} 
Upcycling refers to the process of continual training of pre-trained dense LLMs into SMoE architectures. This approach allows for efficient scaling of model capacity without incurring the substantial computational costs associated with training an SMoE from scratch. The vanilla upcycling~\citep{komatsuzaki2023sparse} approach involves two main steps: 1) replacing specific modules within the dense LLM with SMoE modules by replicating the weights of pre-trained modules as experts and randomly initializing the router networks, and 2) continuously training the resulting SMoE model until convergence or exhaustion of a target compute budget.\par

Initially explored in pre-training to mitigate the training instability and high costs of training SMoE models from scratch~\citep{komatsuzaki2023sparse,he2024upcyclinglargelanguagemodels}, vanilla upcycling has also been effectively adopted in post-training scenarios, resulting in SMoE models with improved generalization performance~\citep{jiang-etal-2025-improved}. Recent research~\citep{sukhbaatar2024branchtrainmix,zhang2024bam} extends upcycling to multi-domain post-training, where the post-training data are divided into well-defined domain subsets. This enables parallel fine-tuning of multiple independent and specialized domain- (or task-) experts. These experts are then merged into a unified SMoE model through uniform weight averaging at non-expert layers, followed by a secondary fine-tuning stage to ensure performance.\par
Despite the simplicity and effectiveness of these approaches, the decision of \emph{where-to-upcycle} throughout upcycling remains heuristically designed. Typically, this involves retaining the FNN~\citep{sukhbaatar2024branchtrainmix} and/or attention layers~\citep{zhang2024bam} as SMoE modules while merging the remaining non-expert parameters through simple unweighted averaging. Such hand-picked strategy risks compromising performance due to architectural biases and uniform averaging of divergent specializations~\citep{li2022branch}.
    
\section{Sparse Interpolated MoE Instruction-Tuning}\label{sec:method}
As outlined in the Introduction, the primary obstacles hindering effective upcycling instruction-tuning of SMoE models revolve around two unresolved challenges: 1) \emph{where-to-upcycle}: determining optimal layer positions in pre-trained models to integrate experts, and 2) \emph{specialization-cooperation trade-off}: balancing expert specialization with inter-expert synergy. Existing methods lack a systematic, data-driven framework to jointly address these challenges. We propose a unified solution grounded in \emph{learnable structured sparsity}, enabling the automatic discovery of upcycling locations in the pre-trained seed LLM while intrinsically managing the specialization-cooperation equilibrium through end-to-end optimization. An overview of our method is shown in Fig.~\ref{fig:overview}.

\subsection{Sparse Interpolated MoE Module}
\subsubsection{Learning \emph{where-to-upcycle} with Structured Sparsity} Current approaches to selecting layers for integrating MoE modules -- such as targeting FNN layers or prioritizing attention components -- rely on architectural heuristics or trial-and-error fine-tuning~\citep{Fedus2021SwitchTS, sukhbaatar2024branchtrainmix}. However, emerging evidence~\citep{zhang2024bam}, along with our own experiments~(Tab.~\ref{tab:ablation-component} (a)), indicates that these fixed architectural biases can lead to suboptimal performance, potentially due to model-specific layer dynamics and data dependencies~\citep{pan2024lisa}. This underscores the need for an adaptive and data-driven method. \par
To address this challenge, we formulate determining  \emph{where-to-upcycle} as a sparse optimization problem. Our key idea is to consider every pre-trained linear layer in the LLM as a potential candidate for upcycling by integrating trainable \emph{per-expert} binary masks $\boldsymbol{z}\in\{0,1\}$ and expert parameters $\boldsymbol{\theta}_i^{\delta}$ into each layer. The linear layer now computes\footnote{The bias term is omitted for neat presentation.}~$\boldsymbol{y}=f_\theta(\boldsymbol{x})=\boldsymbol{\theta}^{\top}\boldsymbol{x},\ \boldsymbol{\theta}=\sum_{i=1}^{M}\alpha_i\cdot \boldsymbol{z}_i\odot \boldsymbol{\theta}_i^{\delta}$, and $M$ is the maximum number of experts allowed to upcycle per mask position. During instruction-tuning, we enforce $L_0$-like sparsity on $\boldsymbol{z}$, allowing the optimization to automatically prune non-essential experts, revealing the learned upcycling strategy.\par

The design of trainable masks $z$ involves a critical trade-off between granularity and computational overhead. While parameter-wise masks enable maximal expressiveness, they are prohibitively computationally expensive for LLMs with billion-scale parameters\footnote{For more details, see the ablation study in Appendix~\ref{app:full_vs_structured_masks}}. To resolve this, we adopt \emph{structured} sparsity on the masks, where masks $\boldsymbol{z} \in \{0,1\}^{\gX}$ gates input neurons in weight matrix $\boldsymbol{\theta} \in \mathbb{R}^{\gY \times \gX}$. This improves scalability while ensuring two key advantages: \textbf{(a)} Fine-grained upcycle control. Masks modulate expert contributions at the granularity of individual neurons, enabling more precise adaptation than layer or FNN upcycling. \textbf{(b)} Hardware-friendly sparsity: structured sparsity patterns align with modern accelerator architectures, avoiding irregular memory access penalties from parameter-wise sparsity.\par 

The implementation still risks overwriting pre-trained knowledge and could incur training instability if all $\boldsymbol{z}_i$, at the same MoE layer, collapse to zeros. As a solution, we anchor the MoEs to the \emph{frozen} pre-trained initialization, $\boldsymbol{\theta}^{\mathtt{pre}}$, reparameterizing the output as
\begin{equation}\label{eq:anchor_moe}    
\boldsymbol{y} = f_{\boldsymbol{\theta}}(\boldsymbol{x}),\ \boldsymbol{\theta}=\boldsymbol{\theta}^{\mathtt{pre}} + \sum_{i=1}^{M}\alpha_i\cdot \boldsymbol{z}_i\odot \boldsymbol{\theta}_i^{\delta}
\end{equation}
This ensures updates remain sparse additions to--rather than replacements of--pre-trained functionality, mitigating catastrophic forgetting while maintaining the same model expressiveness.\par

 One remaining challenge is ensuring differentiability for gradient-based masks optimization. We address this by drawing on the sparse reparameterization method utilized in \citet{schwarz2022metalearning}. At a high-level, each scaler mask $z$ is derived from a latent variable $s$ of a hard concrete distribution~\citep{louizos2018learning} following reparameterization sampling and a transformation, i.e., $z = \!\min(1, \max(0, s)), s\sim q_{\phi}(s)$. Sparsity in $z$ is then enforced via limiting its expected $L_0$-norm in probability, $\bar{L}_0(z)=P(z\neq0)$, which translates to a penalty on the CDF of $q_{\phi}$ as $1-Q_{\phi}(s\leq0)$ that is both differentiable and analytical to compute. This approach avoids non-differentiable thresholding while allowing \emph{exact} zeros in forward passes -- a critical advantage over soft masking. More details are provided in \Cref{app:differentiable_sparsity}. \par
 We enforce sparsity via solving a constrained optimization problem in instruction-tuning, gaining a precise control of the final sparsity in $z$ hence the upcycled experts. More details in \Cref{sec:method:controlled_sparsity}.

\subsubsection{Parameter-Shared Experts with Mask-Driven Specialization}\label{sec:method:specialization}
A fundamental challenge in SMoE upcycling lies in achieving an optimal balance between expert specialization and cooperation. While specialization is crucial for preventing model collapse, effective cooperation among experts is essential for maintaining stable training dynamics. We address this inherent tension through two complementary innovations, leading to our final SIMoE architecture, as illustrated in Fig.~\ref{fig:overview}. \par

First, to promote collaboration among experts, we couple experts by sharing their trainable parameters in the MoE layer in Eqn.~(\ref{eq:anchor_moe}), i.e., $\boldsymbol{\theta}_i^{\delta} = \boldsymbol{\theta}^{\delta}$ for all $i \in [M]$, while maintaining distinct sparsity masks. This design serves dual purposes: it facilitates knowledge transfer through shared parameters and (training) gradients across experts, while substantially reducing memory requirements compared to traditional upcycling approaches that necessitate $M$ separate copies of the original weights.\par

Second, to ensure expert specialization and prevent model collapse due to potential excessive parameter sharing, we impose an orthogonal penalty on the distinct masks, i.e., $\gL^{\mathtt{ortho}}(\boldsymbol{Z})= \|\boldsymbol{Z}\boldsymbol{Z}^\top - \boldsymbol{I} \|_2$ enforcing complementary mask activation patterns while allowing overlap for cooperation. The expression softly penalizes deviation of the dot products between different expert masks -- the off-diagonal terms in $\boldsymbol{Z}\boldsymbol{Z}^\top$ -- from zero, thereby promoting orthogonality. \par

The interplay between parameter sharing and mask orthogonality enables our approach to dynamically discover the optimal balance between expert cooperation and specialization during instruction-tuning. \par


\subsubsection{Instance-level Expert Routing}
Common practices for implementing the router network $h_{\zeta}$ in SMoE systems encompass three primary strategies: token-level, instance-level, and task-level routing. Token-level routing allows different experts to process individual tokens within an input, while instance- and task-level routing maintain consistent expert activation patterns across all tokens in an input or across all inputs within a dataset (or task), respectively. SIMoE remains compatible with all three strategies, as its implementation modifies the configuration of experts only.\par

While task-level routing may incentivize task-specific expert specialization, its applicability is limited by the frequent absence of task meta-information during training and inference. Between token- and instance-level routing, we empirically identify instance-level routing as the optimal choice~(Tab.~\ref{tab:ablation-component} (d)). For the router implementation, each input instruction prompt is processed through the pre-trained LLM to generate token embeddings. We extract the final token embedding as the instance representation, which is then processed by an MLP layer parameterized by $\zeta$ to produce routing logits. These logits are directly transformed through a Softmax function to obtain the expert activations $\boldsymbol{\alpha}$ -- similar to the soft merge mechanism proposed in \citet{Muqeeth2023SoftMO}, which enhances training stability and eliminates the need for auxiliary router losses~\citep{dai-etal-2022-stablemoe}.\par

\subsection{Controlled Compute Cost via Sparsity-constrained Optimization}\label{sec:method:controlled_sparsity}
 While it is straightforward to solve \emph{where-to-upcycle} via a sparsity-regularized training objective, i.e., adding a sparsity regularization term with a fixed coefficient, we instead resort to a more practically compelling approach: enforcing sparsity via solving a sparsity-constrained optimization problem. The latter ensures precise control over the final sparsity in $z$ hence upcycled experts, consequently governs the computational cost of the post-trained LLM at inference -- as expert parameters with zero-valued masks can be safely pruned and eliminated from forward computations. To this end, we consider optimizing the Lagrangian:
\begin{align}\label{eq:training_objective}
\min_{\boldsymbol{\theta^{\delta}}, \boldsymbol{\zeta}, \boldsymbol{\phi}} &\max_{\lambda\geq \mathbf{0}}\ \mathbb{E}_{\mathbf{y},\mathbf{x}}\big[\gL^\mathtt{nll}(\mathbf{y}, \mathbf{x},\boldsymbol{\theta}^{\mathtt{SIMoE}})\big] + \beta \gL^{\mathtt{ortho}}(\boldsymbol{Z}) \nonumber \\
&+ \lambda(\tau - (1-{L}_0(\boldsymbol{Z}))),
\end{align}
where the last term corresponds to the constraint $1 - {L}_0(\boldsymbol{Z}) \geq \tau$ that lower-bounds the sparsity~(number of zeros) in masks, hence upcycled experts, by a hyperparameter $\tau\in[0,1)$.  We adopt simultaneous gradient descent and projected gradient ascent for optimizing the model parameters and the Lagrangian multiplier $\lambda$, respectively. To avoid over-penalizing the model capacity
when surpassing the sparsity constraint, we reset $\lambda$ to zero once the constraint is satisfied~\citep{gallego-posada2022controlled}. More details are deferred to~\Cref{app:optimization}.

\begin{table*}[t!]
\centering
\resizebox{\linewidth}{!}{
\begin{tabular}{lp{1.25cm}lp{1.25cm}p{1.25cm}p{1.25cm}p{1.25cm}p{1.25cm}p{1.25cm}p{1.25cm}p{1.25cm}p{1.25cm}p{1.25cm}p{1.25cm}p{1.25cm}p{1.25cm}p{1.25cm}}
\toprule
\multirow{3}{*}{\textbf{Seed LLM}}&\multirow{3}{*}{\textbf{Method}} & \textbf{Title \newline Gen.} & \textbf{Coref. \newline Res.} & \textbf{Text. \newline Entail.} & \textbf{Quest. \newline Rewrit.} & \textbf{Cause \newline Eff. \newline Class.} & \textbf{Dialog \newline Act \newline Recog.} & \textbf{Ans. \newline Class.} & \textbf{Keyword \newline Tag.} & \textbf{Data to \newline Text} & \textbf{Word \newline Analogy} & \textbf{Overlap \newline Extr.} & \textbf{Grammar \newline Corr.} & \multirow{3}{*}{\textbf{Avg. ($\uparrow$)}} \\
\cmidrule(lr){1-1}\cmidrule(lr){2-2}\cmidrule(lr){3-15}
\multirow{3}{*}{Llama3.2 3B}&Full FT & 40.20 & 55.33 & 58.80 & 67.60 & \textbf{70.52} & 62.38 & \textbf{68.13} & 59.60 & \textbf{52.08} & 39.50 & 66.35 & \textbf{88.68} & 60.76 \\
&Upcyc. & \textbf{41.25} & 57.54 & 62.28 & 67.97 & 68.82 & 66.14 & 67.23 & 63.61 & 51.21 & 46.17 & 62.05 & 87.86 & 61.84 \\
\rowcolor{lightblue}
&  SIMoE & \text{41.14}  & \textbf{57.67} & \textbf{63.17} & \textbf{68.08} & 69.54 & \textbf{68.31} & 67.59 & \textbf{67.87} & 51.40 & \textbf{48.50} & \textbf{68.27} & 87.64 & \textbf{63.26} \\
\cmidrule(lr){1-1}\cmidrule(lr){2-2}\cmidrule(lr){3-15}
\multirow{3}{*}{Llama3 8B}&Full FT & 41.35 & 57.20 & 64.46 & 67.35 & 71.05 & \textbf{73.17} & 67.14 & 66.58 & \textbf{53.04} & 52.71 & 66.93 & \textbf{87.82} & 64.07 \\
&Upcyc. & 41.95 & 62.24 & 64.45 & 68.49 & 73.40 & 69.06 & 66.93 & 66.61 & 52.30 & \textbf{55.55} & 72.05 & 87.59 & 65.05 \\
\rowcolor{lightblue}
&  SIMoE & \textbf{43.04} & \textbf{64.37} & \textbf{66.49} & \textbf{68.86} & \textbf{76.40} & 70.70 & \textbf{68.92} & \textbf{67.79} & 51.73 & \text{52.79} & \textbf{72.06} & 85.38 & \textbf{65.71} \\
\bottomrule
\end{tabular}
}
\caption{\label{tab:fullft_sni}Performance on the SNI benchmark evaluated across 12 unseen task categories, measured in ROUGE-L.}
\end{table*}
\begin{table*}[t!]
\centering
\resizebox{\linewidth}{!}{
\begin{tabular}{lp{1.25cm}p{1.25cm}p{1.25cm}p{1.25cm}p{1.25cm}p{1.25cm}p{1.25cm}p{1.25cm}p{1.25cm}p{1.25cm}p{1.25cm}p{1.25cm}p{1.25cm}}
\toprule
\multirow{2}{*}{\textbf{Method}} & \textbf{MMLU} & \textbf{PopQA} & \textbf{Truthful\newline QA} & \textbf{BBH} & \textbf{DROP} & \textbf{MATH} & \textbf{GSM8K} & \textbf{Human\newline Eval} & \textbf{Human\newline Eval+} & \textbf{IFEval} & \textbf{Alpaca\newline Eval 2} & \textbf{Safety} & \multirow{2}{*}{\textbf{Avg. ($\uparrow$)}} \\
\cmidrule(lr){1-1}\cmidrule(lr){2-14}
$\text{T\"{u}lu}$ v2 8B SFT & 61.8 & 23.3 & 49.4 & 57.1 & \textbf{61.7} & 14.0 & 60.4 &  66.9 & 63.1  & 42.3 &  8.9 & 70.7 & 48.3  \\
RLHFlow v2 SFT & 65.8 &  29.7 & \textbf{56.0} & 69.3 & 57.2 & \textbf{35.7} & \textbf{81.6} & 86.2 & 80.9 & 52.7 & \textbf{13.6} & 43.5 & 56.0  \\
MAmmoTH2 8B &  63.6 &  20.8 & 42.7 & 63.4 &  43.8 &  30.5 & 63.7 & 72.8& 66.4 &  34.9 &  6.5 & 47.8 &  46.4  \\
$\text{T\"{u}lu}$ v3 8B SFT & 65.9 & 29.3 & 46.8 & 67.9 & 61.3 & 31.5 & 76.2 & 86.2 & \textbf{81.4} & 72.8 & 12.4 & 93.1 & 60.4  \\
BTX & 64.5 & \textbf{30.9} & 48.9 & 69.0 & 58.9 & 33.0 & 80.9 & 85.2 & 80.9 & 73.1 & 11.7 & 93.4 & 60.9\\
\rowcolor{lightblue}
SIMoE (Ours) & \textbf{66.5} & 28.7 & 51.6 & \textbf{69.5} & 57.5 & 30.1 & 81.3 & \textbf{86.5} & 81.3 & \textbf{74.1} & 12.4 & \textbf{94.8} & \textbf{61.1}  \\
\bottomrule
\end{tabular}%
}
\caption{\label{tab:full_tulu}Comparison of detailed evaluation results on the $\text{T\"{u}lu}$ 3 eval suite~\citep{lambert2025tulu3pushingfrontiers} between SOTA instruction-tuned modes and ours. All models are instruction-tuned from the pre-trained Llama3.1~8B model.}
\end{table*}

\begin{table}[b!]
\centering
\resizebox{\linewidth}{!}{
\begin{tabular}{lcccccc}
\toprule
&\multicolumn{2}{c}{SNI (Llama3.2 3B)}&\multicolumn{2}{c}{SNI (Llama3 8B)}&\multicolumn{2}{c}{$\text{T\"ulu v3}$ (Llama3.1 8B)}\\
\cmidrule(lr){2-3}\cmidrule(lr){4-5}\cmidrule(lr){6-7}
 Method&\textbf{Params. ($\downarrow$)}
&{\textbf{Avg. ($\uparrow$)}}& \textbf{Params. ($\downarrow$)}
&{\textbf{Avg. ($\uparrow$)}}&\textbf{Params. ($\downarrow$)}
&{\textbf{Avg. ($\uparrow$)}}\\
\cmidrule(lr){1-1}\cmidrule(lr){2-3}\cmidrule(lr){4-5}\cmidrule(lr){6-7}
Upcyc. / BTX&11.67&61.84&30.58&65.05&30.58&60.90\\
\rowcolor{lightblue}
SIMoE (Ours)&\textbf{4.01}&\textbf{63.26}&\textbf{10.04}&\textbf{65.71}&\textbf{10.04}&\textbf{61.10}\\
\bottomrule
\end{tabular}%
}
\caption{\label{tab:combined_up}Overview of average performance and upcycled model capacity on the SNI and $\text{T\"ulu v3}$ benchmarks.}
\end{table}
\section{Experiment}
\subsection{Experimental Setup}
We validate SIMoE on two sets of main experiments that vary in scales of instruction-tuning dataset and model size. For our implementation, we set the maximum number of experts to $M=8$ and the target sparsity constraint to $\tau=75\%$ by default. Thanks to our unique method design~(\Cref{sec:method}), we were able to consider upcycling (attaching SIMoE layer) at every linear layer in the pre-trained LLM, enabling us to optimize for \emph{where-to-upcycle} globally with minimum manual intervention while still maintaining compute feasibility.\par
We compare our results against two types of baselines including: \textbf{(a)} full fine-tuning~(Full FT), and \textbf{(b)} sparse upcycling approaches, including sparse upcycling~\citep{komatsuzaki2023sparse}, and the more empirically competitive BTX~\citep{sukhbaatar2024branchtrainmix} which we use whenever domain labels are available for training. Additional implementation for baselines can be found in~\Cref{app:implementation_details}. We strictly adhere to the official training and evaluation setups of the benchmarks.

\paragraph{Super-NaturalInstructions}
\textbf{SNI}~\citep{wang-etal-2022-super} includes 1,616 instructed NLP tasks over 76 distinct task categories. We strictly follow the official recommended recipe for benchmarking cross-task generalization of instruction-tuned LLM on SNI: training on 64 task categories while leaving out 12 \emph{unseen} categories covering 154 tasks for evaluation only. We adopt ROUGE-L~\citep{lin-2004-ROUGE} for reporting aggregated performance results.

\paragraph{$\text{T\"{u}lu}$-v3} 
The $\text{T\"{u}lu}$-v3 post-training recipe~\citep{lambert2025tulu3pushingfrontiers} provides a large scale instruction-tuning dataset. We use the publicly available SFT data mixture for training, which consists of a total of 939,343 unique training instances from multiple natural and synthetic sub-datasets, spanning a wide range of domains, and the official $\text{T\"{u}lu}$-v3 evaluation suite for testing and reporting performance.

\subsection{Results}

\paragraph{Cross-task generalization}
SIMoE consistently achieves the strongest performance across all of our experiments. As shown in Tab.~\ref{tab:fullft_sni}, SIMoE excels in cross-task generalization on the SNI benchmark, outperforming baselines in at least 7 out of 12 unseen task categories. This results in overall average gains of 2.5\% and 1.6\% over Full FT for the 3B and 8B pre-trained models, respectively.

\paragraph{Scalability and flexibility}
In Tab.~\ref{tab:full_tulu}, SIMoE demonstrates strong generalization performance when transferring to a larger pre-trained model and a relatively larger instruction fine-tuning dataset,i.e., the $\text{T\"{u}lu}$-v3. SIMoE maintains its competitive edge, surpassing all baseline methods on average over 12 common LLM evaluation benchmarks, with a noticeable improvement of 0.6\% over the official $\text{T\"{u}lu}$-v3-8B-SFT model -- the recent open-source state-of-the-art -- while using only 1/3 of the model capacity (number of params) of the BTX upcycled LLM~(Tab.~\ref{tab:combined_up}), demonstrating high efficiency in parameter utilization of our SIMoE module design.\par
Additionally, SIMoE exhibits strong compatibility with different pre-trained LLM architectures: As shown by our earlier experimental results in Tab.~A\ref{tab:sni_t5}, SIMoE continues to outperform Full FT by a significant margin of 2.3\% when switching the pre-trained backbone from Llama3~\citep{grattafiori2024llama3herdmodels} to T5~\citep{raffel2020exploring}, confirming the high flexibility and versatility of our proposed method.

\paragraph{Safety and reliability}
Notably, SIMoE obtains excellent safety evaluation metrics, outperforming $\text{T\"{u}lu}$-v3-8B-SFT by 1.7\% on Safety (Tab.~\ref{tab:full_tulu}), and a remarkable 10\% on the \textit{DoAnythingNow}~\citep{shen2024DoAnythingNow} benchmark in particular (Tab.~A\ref{tab:tulu_safety}, detailed safety results across sub-datasets). The results demonstrate SIMoE's strong resilience against potentially malicious instructions and jailbreak attacks. The dual capability of combining performance improvements with enhanced safety underscores SIMoE's unique potential to mitigate the performance-safety trade-off observed in fine-tuned LLMs~\citep{qi2024finetuning}.


\paragraph{Training and inference cost}
In Fig.~\ref{fig:train_test_compute} and Tab.~A\ref{tab:activated_parameters}, we compare the training and inference compute cost between: (1) SIMoE with a maximum of $M=8$ upcycled sparse interpolated experts at each linear layer, and (2) Sparse upcycling with 4 experts at each FNN block and Top-2 expert routing. Thanks to the proposed learnable, structured sparsity masks in combination with expert parameter sharing, our method significantly reduces model size during training, immediately providing a substantial reduction in peak GPU memory usage. Furthermore, by targeting a final sparsity of $\tau=75\%$ in upcycled experts, our model achieves an smaller inference size, with around $\sim$30\% fewer parameters compared to the number of active parameters per forward-pass in a upcycled SMoE model. 

\subsection{Additional Analysis}
\begin{table}[b!]
\centering
\resizebox{\linewidth}{!}{
\begin{tabular}{lcccccc}
\toprule
\textbf{Model} & \textbf{L.U.} & \textbf{O.P.} & \textbf{S.C.} & \textbf{I.R.} & \textbf{ROUGE-L}$(\uparrow)$ & \textbf{Params.}$(\downarrow)$ (B)\\
\cmidrule(r){1-1} \cmidrule(lr){2-5} \cmidrule(l){6-7}
(a) & & \cmark & \cmark & \cmark & 61.52& 4.01 \\
(b) & \cmark & & \cmark & \cmark & 62.67& 4.01 \\
(c) & & \cmark & & \cmark & 62.54& 6.08\\
(d) & \cmark & \cmark & \cmark & &62.51 &4.01 \\
\cmidrule(r){1-1} \cmidrule(lr){2-5} \cmidrule(l){6-7}
SIMoE & \cmark & \cmark & \cmark & \cmark & \textbf{63.26} & \textbf{4.01} \\
\bottomrule
\end{tabular}%
}
\caption{\label{tab:ablation-component} Ablation results on the SNI benchmark.}
\end{table}
\paragraph{Importance of SIMoE components}
To assess the effectiveness of each component in SIMoE, we conducted an ablation study by systematically removing individual components, resulting in four distinct ablated variants: (a) Instead of \textbf{L}earning \emph{where-to-\textbf{U}pcycle}, we adopt the common practice and upcycle only the FNN layers in the pre-trained LLM for instruction-tuning. (b) We exclude the \textbf{O}rthogonal \textbf{P}enalty on expert masks from our optimization objective. (c) We do not impose \textbf{S}parsity \textbf{C}onstraints on the masks $z$, allowing them to be freely optimized as standard trainable parameters with values on the full real axis. Consequently, upcycling is almost always performed at all linear layers without pruning, hence also no \textbf{L}.\textbf{U}. (d) We replace \textbf{I}nstance-level \textbf{R}outing with token-level routing in the SIMoE module. Other training procedure and non-ablated components remain unchanged.

The variants are compared to the full SIMoE in Table~\ref{tab:ablation-component}. The results clearly demonstrate that the full SIMoE achieves the highest evaluation performance, quantitatively confirming the effectiveness of each proposed component. Notably, enforcing structured sparsity in upcycling not only enhances generalization performance compared to the non-sparse variant (c) -- a result we attribute to its regularization effect -- but also leads to a significant reduction in final model size -- 4.01B vs 6.08B -- which is practically advantageous.

\begin{figure}[t!]
\centering
\includegraphics[width=1.0\linewidth]{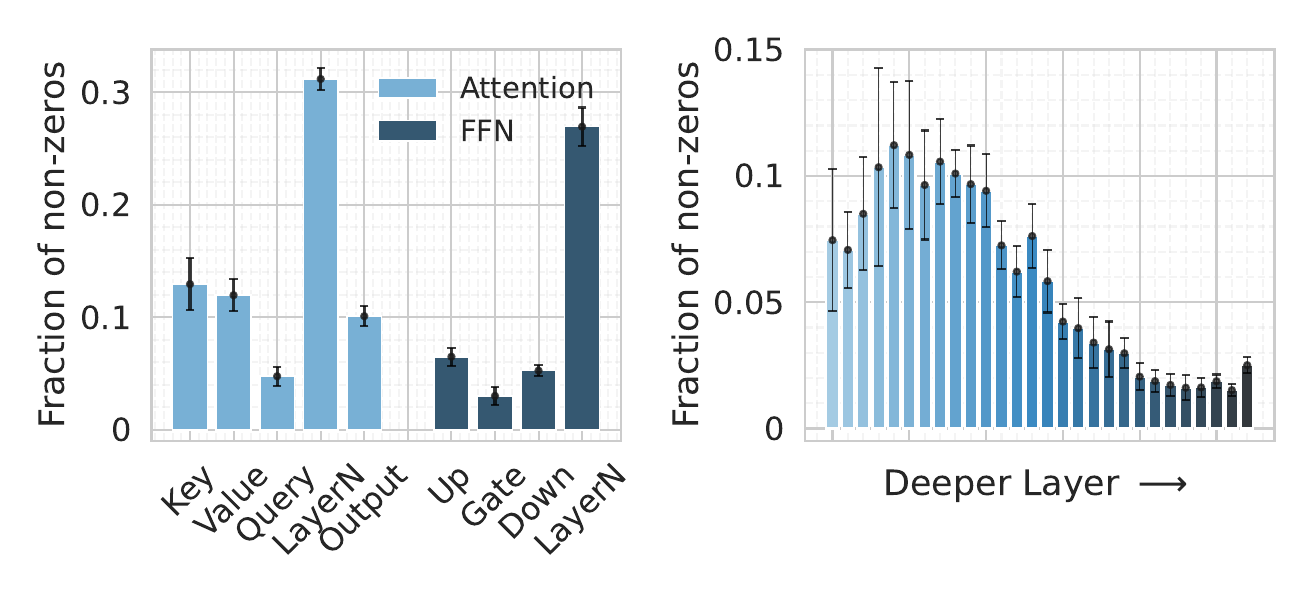}
\caption{Upcycled model capacity after instruction-tuning (i.e., fraction of non-zero expert parameters) grouped by (\emph{left}): layer types, and (\emph{right}): layer depth. The bars represent the average over all experts, and the error bars reflect variation among different experts. }
\label{fig:learned_upcycle}
\end{figure}

\paragraph{Learned sparse upcycling patterns}
In Fig.~\ref{fig:learned_upcycle}, we visualize the distribution of non-zero experts in the upcycled LLM learned by SIMoE from instruction-tuning. Several key observations emerge. \textit{First}, as shown in panel (\emph{right}), upcycling primarily occurs in the shallow and intermediate Transformer layers, with significantly reduced activity in deeper layers. \emph{Second}, panel (\emph{left}) reveals that non-negligible upcycling manifests across all layer types, though with distinct intensity: layer normalization parameters exhibit the highest proportion of upcycled (non-zero) expert parameters, while the gate layer in the FNN demonstrates the lowest. Key, value, and output matrices in the attention block maintain a noticeably higher fraction of non-zero parameters than query weights, aligning with prior work that identified these matrices as crucial for knowledge injection and model editing~\citep{MengLocatingandEditing, gandikota2024unifiedconcepteditingdiffusion}. \par
Notably, the learned upcycling pattern by SIMoE, which achieves the best empirical performance, diverges substantially from manually prescribed strategies (e.g., upcycle FFN only), underscoring the critical advantage of data-driven approaches for determining \emph{where-to-upcycle}.

\paragraph{Interpretable expert routing}
In Fig.~\ref{fig:dendrogram}, we visualize the average expert activation for different tasks. We notice that all experts exhibit some utilization across datasets, and hierarchical clustering of activation similarities reveals a clear dendrogram structure aligned with task and domain relationships. This demonstrates that SIMoE upcycling successfully induces both specialized experts and semantically meaningful routing behaviour.

\begin{figure}[t!]
\centering
\includegraphics[width=1.0\linewidth]{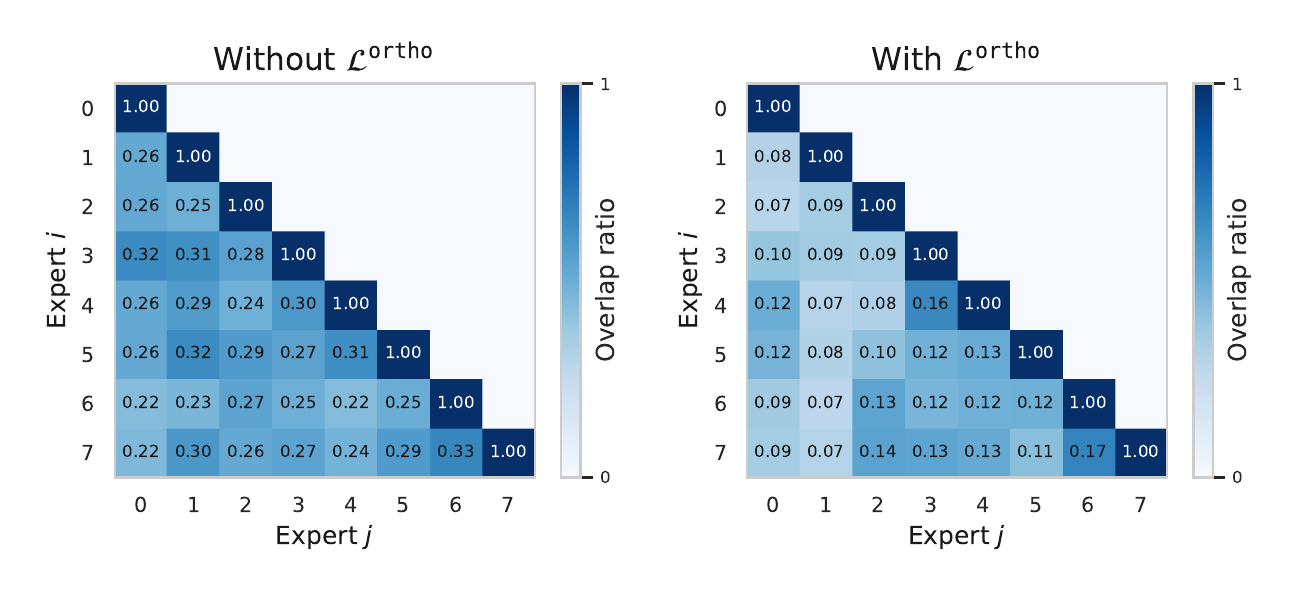}
\caption{Expert overlap ratios of upcycled LLMs post-trained (\textit{left}) without orthogonal penalty, and (\textit{right}) with orthogonal penalty.}
\label{fig:overlaps}
\end{figure}

\paragraph{Specialized experts and orthogonality}
In Fig.~\ref{fig:overlaps}, we assess expert specialization through pairwise mask overlap ratios, defined as $\frac{|(\boldsymbol{z}_i\cap\boldsymbol{z}_j)\neq0|}{|(\boldsymbol{z}_i\cup\boldsymbol{z}_j)\neq0|}$. We observe that experts exhibit higher overlap ratios without the orthogonal penalty; Quantitatively, having the penalty can lead to a noticeable 0.5\% improvement on performance~(Tab.~\ref{tab:ablation-component}). Both observations validate the effectiveness of the penalty. \par
In Fig.~\ref{fig:overlaps}~(\emph{right}), we observe that experts generally have low overlaps -- sharing a small amount of parameters, though domain-similar experts (according to grouping in Fig.~\ref{fig:dendrogram}) exhibit marginally higher overlaps -- for instances, maths- and code-domain experts \{2,6,7\}; general- and safety-domain experts \{3,4\}. The results demonstrate that SIMoE is capable of identifying a balanced shared and expert-specific parameter partitions, enabling nuanced specialization while maintaining strong synergies between distinct experts.

\paragraph{Hyperparameter sensitivity}
\begin{table*}[t!]
\centering
\begin{subtable}{0.45\textwidth}
    \resizebox{\linewidth}{!}{
    \centering
    \begin{tabular}{lcccc}
    \hline
    \textbf{Orthogonality $\beta$} & 0 & $5e^{-6}$ & $\mathbf{5e^{-5}}$ & $5e^{-4}$\\
    \hline
    \textbf{Avg. ROUGE-L} (\textuparrow) & 65.56 & 65.71 & \textbf{65.77} & 65.31 \\
    \hline
    \textbf{Approx. Expert Overlap} \% & 25 & 11 & 7 & 2 \\
    \hline
    \end{tabular}
    }
\end{subtable}%
\hspace{0.05\textwidth} 
\begin{subtable}{0.45\textwidth}
    \resizebox{\linewidth}{!}{
    \centering
    \begin{tabular}{lcccc}
    \hline
    \textbf{Sparsity $\tau$} & 0 & 0.5 & \textbf{0.75} & 0.9 \\
    \hline
    \textbf{Avg. ROUGE-L} (\textuparrow) & 65.37 & 65.31 & \textbf{65.71} & 65.52 \\
    \hline
    \end{tabular}
    }
\end{subtable}
\caption{Impact of key hyperparameters, $\tau$ and $\beta$, on the performance of SIMoE.}
\label{tab:hyperparameter_sensitivity}
\end{table*}

We conduct a comprehensive analysis of the sensitivity of SIMoE to key hyperparameters, specifically the sparsity constraint $\tau$ and the orthogonality penalty coefficient $\beta$. During our experiments, we vary $\beta$ through the values $\{0, 5e^{-6}, 5e^{-5}, 5e^{-4}\}$, and $\tau$ through $\{0, 0.5, 0.75, 0.9\}$, while keeping other hyperparameters constant at their optimal values. We evaluate the trained models on the SNI benchmark using the Llama3 8B model. The results are compared in Tab.~\ref{tab:hyperparameter_sensitivity}.\par

We observe that SIMoE consistently outperforms the best baseline score of 65.05 across all evaluated hyperparameter settings, highlighting its robustness and reliability. Optimal performance is achieved with $\beta\in[5e^{-6},~5e^{-5}]$ and a $\tau$ of 0.75. These settings yield the best empirical results, suggesting that they strike a balance between expert specialization and knowledge transfer. We hypothesize that extreme values of $\beta$ lead to suboptimal outcomes due to excessive or minimal overlap among experts, which either impedes specialization or limits combinatorial generalization capabilities. Similarly, both low and high extremes of $\tau$ result in performance degradation, either through parameter redundancy or excessive sparsity, which constrains model capacity. 
\section{Conclusion}
In this paper, we introduced Sparse Interpolated Mixture-of-Experts (SIMoE) for upcycling dense pre-trained LLMs into SMoEs within a single instruction-tuning stage. By addressing the dual challenges of \emph{where-to-upcycle} and expert specialization-cooperation trade-offs, SIMoE automates the discovery of structurally sparse experts through sparsity-constrained optimization while promoting synergistic yet specialized experts parameter partitions via unique architectural design combined with a orthogonal penalty. Our methods demonstrate empirical superiority and enhanced memory savings compared to existing upcycling instruction-tuning methods, showcasing the efficacy of algorithm-model co-design in unlocking the full potential of upcycling instruction-tuning.\par

\section*{Limitations}
While SIMoE shows promising results, there are two key limitations that warrant attention: (1) Our study is focused solely on NLP tasks, leaving its applicability to multimodal settings (e.g., vision-language models) untested, which is an important area for expanding the framework’s impact; (2) We observe task interference in upcycling baselines and SIMoE, which can negatively affect generalization performance and sometimes cause the MoE models to slightly underperform compared to dense baselines. Future improvements could focus on addressing both challenges.

\bibliography{custom}
\appendix
\section{Method Details}\label{sec:appendix}

\subsection{Differentiable Sparsity}\label{app:differentiable_sparsity}
\citet{louizos2018learning} introduce the hard concrete distribution for modeling sparse gates $z \in [0,1]$. Using the default hyper-parameters from \citep{louizos2018learning}, $\gamma=-0.1, \zeta=1.1, \beta=2/3$, and a random variable $U \sim \text{Unif}(0, 1)$, the hard-concrete distribution models the gate $z$ through:
\begin{align}
&s \sim q_{\phi}(s)= \text{Sigmoid} \left( \frac{1}{\beta} \log \left( \frac{\phi U}{1 - U} \right) \right),\\
&z=\!\min(1, \max(0, s)),
\end{align}
where $q_{\phi}(s)$ is known as the concrete distribution, and $\phi$ is the underlying parameter being optimized. \par

The stochastic nature of $\mathbf{z}$ results in a model that is itself stochastic. Therefore, both its $L_0$-norm and predictions are random quantities. However, as shown in \cite{gallego-posada2022controlled}, $z$ can be replaced with its median $\bar{z}$:
\begin{align}
&\bar{s} =\text{Sigmoid} \left( \frac{\log \phi}{\beta} \right)(\zeta - \gamma)+\gamma),\\
&\bar{z}=\!\min(1, \max(0,\bar{s}),
\end{align}
which is a deterministic function of $\phi$, thus removing the stochastic nature in the gates for training and inference. The stretching and clamping allow the medians to attain values of exactly \emph{0}, producing sparsity in our SIMoE experts when multiplied with expert parameters $\boldsymbol{\theta}^{\delta}$. \par

Furthermore, \citet{louizos2018learning} show that the expected $L_0$-norm of mask $z$ can be expressed in closed-form as:
\begin{align}
\bar{L}_0(z)&=P(z\neq0) = 1- Q_{\phi}(s\leq0)\nonumber\\
&=\text{Sigmoid} \left( \log \phi - \beta \log \frac{-\gamma}{\zeta} \right),
\end{align}
where $Q_{\phi}$ is the CDF of $q_{\phi}$. To this end, sparsity in $z$ is enforced by optimizing $\phi$ to limit $\bar{L}_0(z)$, enabling end-to-end gradient-based optimization under the sparsity constraint in our framework.

\subsection{SIMoE Training Objective~Eqn.~(\ref{eq:training_objective})}\label{app:optimization}
Recall our SIMoE module computes its parameters for inference at each linear layer as:
\begin{align}\label{eq:simoe}
&\boldsymbol{\theta}^{\mathtt{SIMoE}}=\boldsymbol{\theta}^{\mathtt{pre}} + \sum_{i=1}^{M}\alpha_i\cdot \boldsymbol{z}_i\odot \boldsymbol{\theta}^{\delta},
\end{align}
where $\boldsymbol{\theta}^{\mathtt{pre}}$ 
is the frozen pre-trained parameters; $\boldsymbol{\alpha} = h_\zeta(\boldsymbol{x})$ is the expert activation for input prompt $\boldsymbol{x}$ computed by the router network $h_\zeta$; $\{\boldsymbol{z}_i\}^{M}_{i=1}$ and $\boldsymbol{\theta}^{\delta}$ are respectively the sets of distinct masks and shared parameters that jointly define the $M$ experts.

During instruction-tuning, SIMoE aims to uncover \emph{where-to-upcycle} through solving the following sparsity-constrained optimization problem:
\begin{align} \label{eq:sp_con_optimization}
&\min_{\boldsymbol{\theta^{\delta}}, \boldsymbol{\zeta}, \boldsymbol{\phi}}  \mathbb{E}_{\mathbf{y},\mathbf{x}}\big[\gL^\mathtt{nll}(\mathbf{y}, \mathbf{x},\boldsymbol{\theta}^{\mathtt{SIMoE}})\big] +  \beta \gL^{\mathtt{ortho}}(\boldsymbol{Z}),  \nonumber \\
&\ \ \text{s.t.}\quad 1- L_0(\boldsymbol{Z}) \geq \tau,
\end{align}
where the goal is to find model parameters $\{\boldsymbol{\theta}^{\delta}, \zeta, \phi\}$ that minimize a weighted objective between the standard negative log-likelihood loss on the target output $\boldsymbol{y}$ and an orthogonal penalty on the expert masks, while satisfying a constraint that requires mask sparsity, $1-L_0(\boldsymbol{Z})$, to be at least $\tau$. \par

In practice, to enable gradient-based optimization, we solve the Lagrangian associated with the sparsity-constrained optimization problem above. Let $\lambda\geq0$ be the Lagrangian multiplier associated with the sparsity constraint, the min-max Lagrangian problem is then:
\begin{align}
\min_{\boldsymbol{\theta^{\delta}}, \boldsymbol{\zeta}, \boldsymbol{\phi}} &\max_{\lambda\geq \mathbf{0}}\ \mathbb{E}_{\mathbf{y},\mathbf{x}}\big[\gL^\mathtt{nll}(\mathbf{y}, \mathbf{x},\boldsymbol{\theta}^{\mathtt{SIMoE}})\big] + \beta \gL^{\mathtt{ortho}}(\boldsymbol{Z}) \nonumber \\
&+ \lambda(\tau - (1-L_0(\boldsymbol{Z}))),
\end{align}
which mirrors our training objective in Eqn.~(\ref{eq:training_objective}) and we restate it here for completeness.\par
We employ simultaneous gradient descent on model parameters $\{\boldsymbol{\theta}^{\delta}, \zeta, \phi\}$ and projected (to $\mathbb{R}^+$) gradient ascent on $\lambda$, adjusting the strength of the sparse penalty in Eqn.~(\ref{eq:training_objective}) dynamically throughout the optimization~\citep{gallego-posada2022controlled}. The gradient updates in each optimization step are~\footnote{We use vanilla SGD for illustration and omit the learning rates for clarity}:
\begin{align}
&[\boldsymbol{\theta^{\delta}}, \boldsymbol{\zeta}, \boldsymbol{\phi}] \leftarrow \nonumber\\
&[\boldsymbol{\theta^{\delta}}, \boldsymbol{\zeta}, \boldsymbol{\phi}] {\color{red}{-}} \nabla_{[\boldsymbol{\theta^{\delta}}, \boldsymbol{\zeta}, \boldsymbol{\phi}]}\Big[
\mathbb{E}_{\mathbf{y},\mathbf{x}}\big[\gL^\mathtt{nll}(\mathbf{y}, \mathbf{x},\boldsymbol{\theta}^{\mathtt{SIMoE}})\nonumber\\
&+ \beta \gL^{\mathtt{ortho}}(\boldsymbol{Z}) - \lambda(1-L_0(\boldsymbol{Z}))\Big], \label{eq:primary_update}\\
&[\lambda] \leftarrow \max\Big(0, [\lambda] {\color{red}{+}} (\tau - (1-L_0(\boldsymbol{Z}))) \Big).
\end{align}
The sparsity $(1-L_0(\boldsymbol{Z}))$ in Eqn.~(\ref{eq:primary_update}) remains non-differentiable w.r.t. $\phi$, the parameters of the hard-concrete distribution producing masks $\boldsymbol{Z}$. Thus, we compute the expected sparsity $(1-\bar{L}_0(\boldsymbol{Z}))$, which is differentiable w.r.t. $\phi$ (see~\Cref{app:differentiable_sparsity}).\par
Overall, $\lambda$ continues to increase, progressively emphasizing the sparsity penalty $(1-\bar{L}_0(\boldsymbol{Z}))$ until the constraint is satisfied during optimization.

\section{Experiment Details}
\subsection{Implementation}~\label{app:implementation_details}
For our method, we set the maximum number of experts to $M=8$ and the target sparsity constraint to $\tau=75\%$ by default, unless stated otherwise. We initialize the trainable parameters so that the SIMoE model $\theta^\texttt{SIMoE}$ in Eqn.~(\ref{eq:simoe}) behaves identically the pre-trained LLM $\theta^\mathtt{pre}$ at the start of instruction-tuning. Specifically, we initialize the shared expert parameters, $\theta^{\delta}$, to zero, and the mask parameters $\phi$ from a Gaussian distribution with a mean that results in an initial expected sparsity of 0.05 in each mask, i.e., $P(z=0)=Q_{\phi}(s\leq0)=0.05,\forall z\in \boldsymbol{Z}$. Thanks to our unique method design~(\Cref{sec:method}), we are able to consider upcycling (attaching SIMoE layer) at every linear layer in the pre-trained LLM. This enables us to optimize \emph{where-to-upcycle} globally with minimal manual intervention while still maintaining compute feasibility. The detailed training hyperparameters for our method can be found in Tab.~\ref{tab: hyperparameters} below. \par
\begin{table}[h]
\centering
\begin{tabular}{lcc}
\hline
\textbf{Hyperparameter} & \textbf{SNI} & \textbf{{$\text{T\"{u}lu}$-v3}} \\
\cmidrule(r){1-1} \cmidrule(lr){2-3} 
Learning rate & 2e-5 & 2e-5 \\
Learning rate scheduler& constant & linear\\
Batch size & 16 & 128 \\
Optimizer & Adam & Adam  \\
Sparsity ($\tau$)& 0.75& 0.75\\
Orthogonality ($\beta$) &5e-6 &5e-6 \\
Experts ($M$)& 8 & 8 \\
\hline
\end{tabular}
\caption{Training hyperparameters used for SIMoE in SNI and $\text{T\"{u}lu}$-v3 experiments.}
\label{tab: hyperparameters}
\end{table}

For the baseline sparse upcycling and BTX, following the original implementation described in~\citep{sukhbaatar2024branchtrainmix}, we upcycle FNN experts in the pre-trained dense LLM, and use a default Top-2 routing function. We use a load-balancing loss with coefficient $1e-2$~\citep{shazeer2017,Fedus2021SwitchTS} and a router-$z$ loss with strength $1e-3$ to stabilize training~\citep{dai-etal-2022-stablemoe}. We note that both methods can eventually produce an SMoE model with the same architecture and model size, given the same number of upcycled experts and the same seed LLM as fixed hyperparameters.\par
On the Super-NaturalInstruction benchmark, we set the number of upcycling experts to 4, which already yields a model size approximately equals to 4 times that of the original LLM. For Tulu-v3 experiment with BTX~\citep{sukhbaatar2024branchtrainmix}, we initialize experts with FNN from independently trained domain experts on pre-defined domains in the Tulu-v3 training data mixture, including: math, code, safety, instruction following, multilingual, knowledge recall and general. We also include the pre-trained checkpoint as an additional expert as done in~\citet{sukhbaatar2024branchtrainmix}.

\section{Additional Results}

\subsection{Full Granular vs. Structured Sparsity Masks}\label{app:full_vs_structured_masks}

We explore the impact of mask granularity on the trade-off between model performance and computational efficiency, focusing on SIMoE models.

Full granular masks, while potentially enhancing model expressiveness by applying distinct masks to each parameter, present significant challenges. For models scaled to a billion parameters with $M$ experts, these masks can demand up to $M$ times the original model size, making them \textbf{less scalable} with larger models. Additionally, training with parameter-wise sparse, \textit{stochastic} masks can lead to \textbf{optimization difficulties}, such as training instability and an increased risk of overfitting.

To assess these effects, we conducted experiments with the Llama3.2 1B model on the SNI benchmark. As shown in Table~\ref{tab:mask_comparison}, the results demonstrate that structured sparsity masks not only achieve superior performance but also offer better computational efficiency. This highlights the advantages of adopting structured sparsity in large-scale models.

\begin{table}[ht]
\centering
\begin{tabular}{lcc}
\hline
\textbf{Mask Type} & \textbf{Rouge-L (\textuparrow)} & \textbf{Param.(\textdownarrow)} \\
\hline
Full granular & 59.36 & 3.42B \\ Structured & \textbf{59.94} & \textbf{1.45B} \\
\hline
\end{tabular}
\caption{Comparison of SIMoE with full granular masks and structured sparsity masks.}
\label{tab:mask_comparison}
\end{table}

\subsection{Activated Parameter Counts in Masked Models}
Tab.~\ref{tab:activated_parameters} presents a detailed comparison of activated versus total parameter counts for SMoE (Upcyc/BTX) and our proposed SIMoE model. Please refer to Fig.~\ref{fig:train_test_compute} for a qualitative visualization of these results. 
\begin{table}[ht]
\centering
\begin{tabular}{lcc}
\hline
\textbf{Method} & \textbf{3B Seed} & \textbf{8B Seed} \\
\hline
Upcyc/BTX & 7 / 11.67B & 13.64 / 30.58B \\
SIMoE (Ours) & 4.01 / 4.01B & 10.4 / 10.4B \\
\hline
\end{tabular}
\caption{Comparison of activated versus total parameter counts for SMoE and SIMoE models.}
\label{tab:activated_parameters}
\end{table}

\subsection{Detailed $\text{T\"ulu-v3}$ Safety Evaluation Results}
\begin{table}[h!]
\centering
\begin{tabular}{lcc}
\hline
\textbf{Benchmark} & \textbf{Tulu 3 8B SFT} & \textbf{Ours} \\
\cmidrule(r){1-1} \cmidrule(lr){2-3} 
HarmBench & \textbf{98.4} & 97.5 \\
XSTest & 90.4 & 90.4 \\
WildGuardTest & 99.2 & \textbf{99.5} \\
Jailbreaktrigger & 95.8 & 95.8 \\
DoAnythingNow & 88.3 & \textbf{98.0} \\
WildjailbreakTest & 86.7 & \textbf{87.7} \\
\cmidrule(r){1-1} \cmidrule(lr){2-3} 
\textbf{Average $(\uparrow)$} & 93.1 & \textbf{94.8} \\
\hline
\end{tabular}
\caption{\label{tab:tulu_safety}Breakdown of safety scores by benchmark of ours compared with the open-source state-of-the-art, Tulu 3 8B SFT~\citep{lambert2025tulu3pushingfrontiers}.}
\end{table}
Tab.~\ref{tab:tulu_safety} presents a comparison of safety scores between 
SIMoE and the Tulu 3 8B SFT~\citep{lambert2025tulu3pushingfrontiers}, across various benchmarks. While Tulu 3 8B SFT outperforms in HarmBench, our model maintains a competitive edge overall, with an average score of 94.8 compared to Tulu's 93.1. This highlights the potential of our approach in safety-critical evaluations.

\subsection{Instruction-tuned T5-XL on Super-NaturalInstruction}
\begin{table}[h]
\centering
\begin{tabular}{lcc}
\toprule
\textbf{Tasks} & \textbf{Tk-Instruct 3B} & \textbf{Ours} \\
\cmidrule(r){1-1} \cmidrule(lr){2-3} 
\cmidrule(r){1-1} \cmidrule(lr){2-3} 
\textbf{Average} $(\uparrow)$ &  56.72& 59.21 \\
\bottomrule
\end{tabular}
\caption{\label{tab:sni_t5}Performance of Tk-instruct~\citep{wang-etal-2022-super} and our method on the SNI benchmark. Both models start from pre-trained T5-XL~\citep{raffel2020exploring}.}

\end{table}

\subsection{Detailed $\text{T\"ulu-v3}$ and SNI Evaluation Results}
We present detailed evaluation results, including parameter counts, comparing SIMoE with previous sparse upcycling methods in Tables~\ref{tab:llama_sni} and~\ref{tab:llama_tulu} below.

\begin{table*}[t!]
\centering
\resizebox{\linewidth}{!}{
\begin{tabular}{lp{1.25cm}p{1.25cm}p{1.25cm}p{1.25cm}p{1.25cm}p{1.25cm}p{1.25cm}p{1.25cm}p{1.25cm}p{1.25cm}p{1.25cm}p{1.25cm}p{1.25cm}p{1.25cm}}
\toprule
\multirow{3}{*}{\textbf{Method}} &\textbf{Params.\newline (B) ($\downarrow$)} & \textbf{Title \newline Gen.} & \textbf{Coref. \newline Res.} & \textbf{Text. \newline Entail.} & \textbf{Quest. \newline Rewrit.} & \textbf{Cause \newline Eff. \newline Class.} & \textbf{Dialog \newline Act \newline Recog.} & \textbf{Ans. \newline Class.} & \textbf{Keyword \newline Tag.} & \textbf{Data to \newline Text} & \textbf{Word \newline Analogy} & \textbf{Overlap \newline Extr.} & \textbf{Grammar \newline Corr.} & \multirow{3}{*}{\textbf{Avg. ($\uparrow$)}} \\
\cmidrule(lr){1-1}\cmidrule(lr){2-2} \cmidrule(lr){3-14}\cmidrule(lr){15-15}
Upcycling (3B) & 11.67 & \textbf{41.25} & 57.54 & 62.28 & 67.97 & 68.82 & 66.14 & 67.23 & 63.61 & 51.21 & 46.17 & 62.05 & \textbf{87.86} & 61.84 \\
\rowcolor{lightblue}
SIMoE (3B) & \textbf{4.01} & 41.14 & \textbf{57.67} & \textbf{63.17} & \textbf{68.08} & \textbf{69.54} & \textbf{68.31} & \textbf{67.59} & \textbf{67.87} & \textbf{51.40} & \textbf{48.50} & \textbf{68.27} & 87.64 & \textbf{63.26} \\
\cmidrule(lr){1-1}\cmidrule(lr){2-2} \cmidrule(lr){3-14}\cmidrule(lr){15-15}
Upcycling (8B) &30.58 & 41.95 & 62.24 & 64.45 & 68.49 & 73.40 & 69.06 & 66.93 & 66.61 & \textbf{52.30} & \textbf{55.55} & 72.05 & \textbf{87.59} & 65.05 \\
\rowcolor{lightblue}
SIMoE (8B) & \textbf{10.04} & \textbf{43.04} & \textbf{64.37} & \textbf{66.49} & \textbf{68.86} & \textbf{76.40} & \textbf{70.70} & \textbf{68.92} & \textbf{67.79} & 51.73 & 52.79 & \textbf{72.06} & 85.38 & \textbf{65.71} \\
\bottomrule
\end{tabular}%
}
\caption{\label{tab:llama_sni}Performance on the SNI benchmark evaluated across 12 unseen task categories, measured in ROUGE-L. The top and bottom sections respectively show results for the Llama3.2-3B and the Llama3-8B pre-trained models.}
\end{table*}
\begin{table*}[t!]
\centering
\resizebox{\linewidth}{!}{
\begin{tabular}{lp{1.25cm}p{1.25cm}p{1.25cm}p{1.25cm}p{1.25cm}p{1.25cm}p{1.25cm}p{1.25cm}p{1.25cm}p{1.25cm}p{1.25cm}p{1.25cm}p{1.25cm}p{1.25cm}}
\toprule
\multirow{2}{*}{\textbf{Method}} &\textbf{Params.\newline (B) ($\downarrow$)} & \textbf{MMLU} & \textbf{PopQA} & \textbf{Truthful\newline QA} & \textbf{BBH} & \textbf{DROP} & \textbf{MATH} & \textbf{GSM8K} & \textbf{Human\newline Eval} & \textbf{Human\newline Eval+} & \textbf{IFEval} & \textbf{Alpaca\newline Eval 2} & \textbf{Safety} & \multirow{2}{*}{\textbf{Avg. ($\uparrow$)}} \\
\cmidrule(lr){1-1}\cmidrule(lr){2-2} \cmidrule(lr){3-14}\cmidrule(lr){15-15}
BTX &30.58 &64.5 & \textbf{30.9} & 48.9 & 69.0 & \textbf{58.9} & \textbf{33.0} & 80.9 & 85.2 & 80.9 & 73.1 & 11.7 & 93.4 & 60.9\\
\rowcolor{lightblue}
SIMoE & \textbf{10.04}  &\textbf{66.5} & 28.7 & \textbf{51.6} & \textbf{69.5} & 57.5 & 30.1 & \textbf{81.3} & \textbf{86.5} & \textbf{81.3} & \textbf{74.1} & \textbf{12.4} & \textbf{94.8} & \textbf{61.1}  \\
\bottomrule
\end{tabular}%
}
\caption{\label{tab:llama_tulu}Overview of results on the $\text{T\"{u}lu}$ 3 eval suite~\citep{lambert2025tulu3pushingfrontiers}. All models are instruction-tuned from the pre-trained Llama3.1~8B model.}
\end{table*}

\end{document}